\pdfoutput=1

\documentclass[11pt]{article}

\usepackage[]{acl}

\usepackage{times}
\usepackage{latexsym}

\usepackage[T1]{fontenc}
\usepackage[utf8]{inputenc}

\usepackage{microtype}
\usepackage{amsmath}
\usepackage{algorithm}
\usepackage{algpseudocode}
\usepackage[capitalise,noabbrev]{cleveref}

\usepackage{tikz}
\usetikzlibrary{calc}
\usetikzlibrary{positioning,arrows,automata}
\usetikzlibrary{shapes.geometric}
\usepackage{pgfplots}
\usepgfplotslibrary{groupplots}
\usepackage{subcaption}
\usepackage{amssymb}
\let\emptyset\varnothing
\usepackage{hyperref}
\pgfplotsset{compat/show suggested version=false}

\usepackage{booktabs}
\usepackage{multirow}
\usepackage{graphicx}

\algrenewcommand{\Return}{\State\algorithmicreturn~}

\usepackage[normalem]{ulem} %

\DeclareMathOperator{\prefix}{prefix}
\DeclareMathOperator{\LCP}{LCP}

\title{CUNI-KIT System for Simultaneous Speech Translation Task
at IWSLT 2022
}

\author{
    Peter Pol\'ak\textsuperscript{1} \and 
    Ngoc-Quan Ngoc\textsuperscript{2} \and 
    Tuan-Nam Nguyen\textsuperscript{2} \and 
    Danni Liu\textsuperscript{3}  \AND
    Carlos Mullov\textsuperscript{2} \and
    Jan Niehues\textsuperscript{2} \and 
    Ond\v{r}ej Bojar\textsuperscript{1} \and 
    Alexander Waibel\textsuperscript{2,4} \\
    \texttt{polak@ufal.mff.cuni.cz} \\
    \textsuperscript{1} Charles University \\
    \textsuperscript{2} Karlsruhe Institute of Technology \\
    \textsuperscript{3} Maastricht University \\
    \textsuperscript{4} Carnegie Mellon University \\
    }

\begin{document}
\maketitle
\begin{abstract}
In this paper, we describe our submission to the Simultaneous Speech Translation at IWSLT 2022. We explore strategies to utilize an offline model in a simultaneous setting without the need to modify the original model. In our experiments, we show that our onlinization algorithm is almost on par with the offline setting while being $3\times$ faster than offline in terms of latency on the test set. We also show that the onlinized offline model outperforms the best IWSLT2021 simultaneous system in medium and high latency regimes and is almost on par in the low latency regime. We make our system publicly available.\footnote{\url{https://hub.docker.com/repository/docker/polape7/cuni-kit-simultaneous}}
\end{abstract}

\section{Introduction}

This paper describes the CUNI-KIT submission to the Simultaneous Speech Translation task at IWSLT 2022 \citep{iwslt:2022} by Charles University (CUNI) and Karlsruhe Institute of Technology (KIT). 

Recent work on end-to-end (E2E) simultaneous speech-to-text translation (ST) is focused on training specialized models specifically for this task. The disadvantage is the need of storing an extra model, usually a more difficult training and inference setup, increased computational complexity \citep{han-etal-2020-end,liu-etal-2021-ustc} and risk of performance degradation if used in offline setting \citep{Liu2020}.

In this work, we base our system on a robust multilingual offline ST model that leverages pretrained wav2vec 2.0 \citep{baevski2020wav2vec} and mBART \citep{liu-etal-2020-multilingual-denoising}. We revise the onlinization approach by \citet{Liu2020} and propose an improved technique with a fully controllable quality-latency trade-off. We demonstrate that without any change to the offline model, our simultaneous system in the mid- and high-latency regimes is on par with the offline performance. At the same time, the model outperforms previous IWSLT systems in medium and high latency regimes and is almost on par in the low latency regime. Finally, we observe a problematic behavior of the average lagging metric for speech translation \citep{ma2020simuleval} when dealing with long hypotheses, resulting in negative values. We propose a minor change to the metric formula to prevent this behavior.

Our contribution is as follows:
\begin{itemize}
    \item We revise and generalize onlinization proposed by \citet{Liu2020,nguyen21c_interspeech} and discover parameter enabling quality-latency trade-off,
    \item We demonstrate that one multilingual offline model can serve as simultaneous ST for three language pairs,
    \item We demonstrate that an improvement in the offline model leads also to an improvement in the online regime,
    \item We propose a change to the average lagging metric that avoids negative values.
\end{itemize}

\section{Related Work}
Simultaneous speech translation can be implemented either as a (hybrid) cascaded system \citep{kolss2008stream,niehues16_interspeech,elbayad-etal-2020-trac,Liu2020,bahar-etal-2021-without} or an end-to-end model \citep{han-etal-2020-end,liu-etal-2021-ustc}. Unlike for the offline speech translation where cascade seems to have the best quality, the end-to-end speech translation offers a better quality-latency trade-off \citep{ansari-etal-2020-findings,liu-etal-2021-ustc,anastasopoulos-etal-2021-findings}. 

End-to-end systems use different techniques to perform simultaneous speech translation. \citet{han-etal-2020-end} uses wait-$k$ \citep{ma2019stacl} model and metalearning \citep{9054759} to alleviate the data scarcity. \citet{Liu2020} uses a unidirectional encoder with monotonic cross-attention to limit the dependence on future context. Other work \citep{liu-etal-2021-ustc} proposes Cross Attention augmented Transducer (CAAT) as an extension of RNN-T \citep{graves2012sequence}.

\citet{nguyen21c_interspeech} proposed a hypothesis stability detection for automatic speech recognition (ASR). The \emph{shared prefix} strategy finds the longest common prefix in all beams. \citet{Liu2020} explore such strategies in the context of speech recognition and translation. The most promising is the longest common prefix of two consecutive chunks. The downside of this approach is the inability to parametrize the quality-latency trade-off. We directly address this in our work. %

\section{Onlinization}
In this section, we describe the onlinization of the offline model and propose two ways to control the quality-latency trade-off.

\subsection{Incremental Decoding}

Depending on the language pair, translation tasks may require reordering or a piece of information that might not be apparent until the source utterance ends. In the offline setting, the model processes the whole utterance at once, rendering the strategy most optimal in terms of quality. If applied in online mode, this ultimately leads to a large latency. One approach to reducing the latency is to break the source utterance into chunks and perform the translation on each chunk. 

In this paper, we follow the incremental decoding framework described by \citet{Liu2020}. We break the input utterance into small fixed-size chunks and decode each time after we receive a new chunk. After each decoding step, we identify a stable part of the hypothesis using \emph{stable hypothesis detection}. The stable part is sent to the user (``committed'' in the following) and is no longer changed afterward (i.e., no retranslation).\footnote{This is a requirement for the evaluation in the Simultaneous Speech Translation task at IWSLT 2022.} %
Our current implementation assumes that the whole speech input fits into memory, in other words, we are only adding new chunks as they are arriving. This simplification is possible because the evaluation of the shared task is performed on segmented input, on individual utterances.
With each newly arrived input chunk, the decoding starts with forced decoding of the already committed tokens and continues with beam search decoding.

\subsection{Chunk Size}
Speech recognition and translation use chunking for simultaneous inference with various chunk sizes ranging from 300 ms to 2 seconds \citep{liu2020low,nguyen21c_interspeech} although the literature suggests that the turn-taking in conversational speech is shorter, around 200 ms \citep{levinson2015timing}. We investigate different chunk sizes in combination with various stable hypothesis detection strategies. As we document later, the chunk size is the principal factor that controls the quality-latency trade-off.

\subsection{Stable Hypothesis Detection}
Committing hypotheses from incomplete input %
presents a possible risk of introducing errors. To reduce the instability and trade time for quality, we employ a \emph{stable hypothesis detection}. Formally, we define a function $prefix(W)$ that, given a set of hypotheses (i.e., $W_{all}^c$ if we want to consider the whole beam or $W_{best}^c$ for the single best hypothesis obtained during the beam search decoding of the $c$-th chunk), outputs a stable prefix. We investigate several functions:

\paragraph{\textbf{Hold-$n$}} \citep{Liu2020} Hold-$n$ strategy selects the best hypothesis in the beam
and deletes the last $n$ tokens from it:
\begin{equation}
    \prefix(W_{best}^c) = W_{0:\max(0, |W|-n)},
\end{equation}
where $W_{best}^c$ is the best hypothesis obtained in the beam search of $c$-th chunk. If the hypothesis has only $n$ or fewer tokens, we return an empty string.

\paragraph{LA-$n$} Local agreement \citep{Liu2020} displays the agreeing prefixes of the two consecutive chunks. Unlike the hold-$n$ strategy, the local agreement does not offer any explicit quality-latency trade-off. We generalize the strategy to take the agreeing prefixes of $n$ consecutive chunks. 

During the first $n-1$ chunks, we do not output any tokens. From the $n$-th chunk on, we identify the longest common prefix of the best hypothesis of the $n$ consecutive chunks:
\begin{multline}
        \prefix(W_{best}^c) =  \\
    \begin{cases}
    \emptyset, & \text{if $c < n$},\\
    \LCP(W_{best}^{c-n+1}, ..., W_{best}^{c}), & \text{otherwise},
    \end{cases}
\end{multline}
where $LCP(\cdot)$ is longest common prefix of the arguments.

\paragraph{SP-$n$} Shared prefix \citep{nguyen21c_interspeech} strategy displays the longest common prefix of all the items in the beam of a chunk. Similarly to the LA-$n$ strategy, we propose a generalization to the longest common prefix of all items in the beams of the $n$ consecutive chunks:

\begin{multline}
        \prefix(W_{all}^c) =  \\
    \begin{cases}
    \emptyset, & \text{if $c < n$},\\
    \LCP(W_{\text{beam $1...B$}}^{c-n+1}, ..., W_{\text{beam $1...B$}}^c), & \text{otherwise},
    \end{cases}
\end{multline}
i.e., all beam hypotheses $1,...,B$ (where $B$ is the beam size) of all chunks $c-n+1,...,c$. 

\subsection{Initial Wait} 
The limited context of the early chunks might result in an unstable hypothesis and an emission of erroneous tokens. The autoregressive nature of the model might cause further performance degradation in later chunks. One possible solution is to use longer chunks, but it inevitably leads to a higher latency throughout the whole utterance. To mitigate this issue, we explore a lengthening of the first chunk. We call this strategy an initial wait.

\section{Experiments Setup}
In this section, we describe the onlinization experiments.

\subsection{Evaluation Setup}
We use the SimulEval toolkit \citep{ma2020simuleval}. The toolkit provides a simple interface for evaluation of simultaneous (speech) translation. It reports the quality metric BLEU \citep{papineni2002bleu,post2018call} and latency metrics Average Proportion (AP, \citealt{cho2016can}), Average Lagging (AL, \citealt{ma2019stacl}), and Differentiable Average Lagging (DAL, \citealt{cherry2019thinking}) modified for speech source.  

Specifically, we implement an \texttt{Agent} class. We have to implement two important functions: \texttt{policy(state)} and \texttt{predict(state)}, where \texttt{state} is the state of the agent (e.g., read processed input, emitted tokens, ...). The policy function returns the action of the agent: (1) \texttt{READ} to request more input, (2) \texttt{WRITE} to emit new hypothesis tokens. 

We implement the \texttt{policy} as specified in \cref{alg:policy}. The default action is \texttt{READ}. If there is a new chunk, we perform the inference and use the $prefix(W^c)$ function to find the stable prefix. If there are new tokens to display (i.e., $|prefix(W^c)| > |prefix(W^{c-1})|$), we return the  \texttt{WRITE} action. As soon as our agent emits an end-of-sequence (EOS) token, the inference of the utterance is finished by the SimulEval. We noticed that our model was emitting the EOS token quite often, especially in the early chunks. Hence, we ignore the EOS if returned by our model and continue the inference until the end of the source.\footnote{This might cause an unnecessary increase in latency, but it might be partially prevented by voice activity detection.}%

\begin{algorithm}
\caption{Policy function}\label{alg:policy}
\begin{algorithmic}
\Require $state$

\If{$state$.$new\_input > chunk\_size$ }
    \State $hypothesis \gets predict(state)$
    \If{$|hypothesis| > 0$}
        \Return $WRITE$
    \EndIf
\EndIf
\Return $READ$
\end{algorithmic}
\end{algorithm}

\subsection{Speech Translation Models}
In our experiments, we use two different models. First, we do experiments with a monolingual \emph{Model A}, then for the submission, we use a multilingual and more robust \emph{Model B}.\footnote{We also did experiments with a dedicated English-German model similar to Model B (i.e., based on wav2vec and mBART), but it performed worse both in offline and online setting compared to the multilingual version.}

Model A is the KIT IWSLT 2020 model for the Offline Speech Translation task. Specifically, it is an end-to-end English to German Transformer model with relative attention. For more described description, refer to \citet{pham2020kit}.

\subsubsection{Multilingual Model}
\label{sec:models}
For the submission, we use a multilingual Model B. We construct the SLT architecture with the encoder based on the wav2vec 2.0~\cite{baevski2020wav2vec} and the decoder based on the autoregressive language model pretrained with mBART50~\cite{tang2020multilingual}.

\paragraph{wav2vec 2.0} is a Transformer encoder model which receives raw waveforms as input and generates high-level representations. The architecture consists of two main components: first, a convolution-based \textit{feature extractor} downsamples long audio waveforms into features that have similar lengths with spectrograms.%
After that, a deep Transformer encoder uses self-attention and feed-forward neural network blocks to transform the features without further downsampling. 

During the self-supervised training process, the network is trained with a contrastive learning strategy~\cite{baevski2020wav2vec}, in which the already downsampled features are randomly masked and the model learns to predict the quantized latent representation of the masked time step.%

During the supervised learning step, we freeze the feature extraction weights to save memory since the first layers are among the largest ones. We fine-tune all of the weights in the Transformer encoder. Moreover, to make the model more robust to the fluctuation in absolute positions and durations when it comes to audio signals,%
we added the relative position encodings~\cite{dai-etal-2019-transformer, pham2020relative} to alleviate this problem.\footnote{This has the added advantage of better generalization in situations where training and testing data are segmented differently.}

Here we used the same pretrained model with the speech recognizer, with the large architecture pretrained with $53k$ hours of unlabeled data. 

\paragraph{mBART50} is an encoder-decoder Transformer-based language model. During training, instead of the typical language modeling setting of predicting the next word in the sequence, this model is trained to reconstruct a sequence from its noisy version~\cite{lewis2019bart} and later extended to a multilingual version~\cite{liu-etal-2020-multilingual-denoising,tang2020multilingual} in which the corpora from multiple languages are combined during training. mBART50 is the version that is pretrained on $50$ languages.

The mBART50 model follows the Transformer encoder and decoder~\cite{vaswani2017attention}. During fine-tuning, we combine the mBART50 decoder with the wav2vec 2.0 encoder, where both encoder and decoder know one modality. The cross-attention layers connecting the decoder with the encoder are the parts that require extensive fine-tuning in this case, due to the modality mismatch between pretraining and fine-tuning.

Finally, we use the model in a multilingual setting, i.e., for English to Chinese, German, and Japanese language pairs by training on the combination of the datasets. The mBART50 vocabulary contains language tokens for all three languages and can be used to control the language output~\cite{Ha2016}.

For more details on the model refer to \citet{kit:2022}.

\subsection{Test Data}
For the onlinization experiments, we use MuST-C \citep{CATTONI2021101155} \texttt{tst-COMMON} from the v2.0 release. We conduct all the experiments on the English-German language pair. 

\section{Experiments and Results}
In this section, we describe the experiments and discuss the results. %

\subsection{Chunks Size}
\label{exp:chunks}
\begin{figure}[ht]
	\centering
	\footnotesize
	\begin{tikzpicture}
	\pgfplotsset{
		scaled y ticks = false,
		width=0.99\columnwidth,
		height=0.6\columnwidth,
		axis on top,
		grid=major,
		xlabel style={at={(0.5,0.1)}},
		extra x tick style={grid=none, draw=none, tick label style={xshift=0cm,yshift=.30cm},tick style={draw=none}},
		extra x tick label={$\mathbin{/\mkern-6mu/}$},
		ylabel shift={-1.5em},
		ylabel style={align=center, at={(0.1,0.5)}}
%		try min ticks=10
	}
	\begin{groupplot}[ 
	group style={
		group size=1 by 1,
		vertical sep=0pt,
		horizontal sep=0pt
	},
	]
	% ----------
	% Plot [0, 0]
	%-----------
	\nextgroupplot[
	xlabel={Average Lagging (seconds)},
	ylabel={BLEU},
    x label style={at={(axis description cs:0.5,0.02)},anchor=north},
    y label style={at={(axis description cs:0.11,.5)},anchor=south},
	xticklabels={0,0.5,1,1.5,2,2.5,3}, % Fake here the x-value for the offline thing
	xtick={0,0.5,1,1.5,2,2.5,3},
	yticklabels={15,20,25,30},
	ytick={15,20,25,30},
	xmin=-0.1,
	xmax=3.1,
	ymin=16, %13.9,
	ymax=32,
	legend style={legend columns=1,fill=white,draw=black,anchor=west,align=left, font=\footnotesize}, %at={($(0,0)+(1cm,1cm)$)},
	legend to name=legend_sizes
	]
	\addplot+[sharp plot, solid, mark size=1.3pt, ultra thick] table {
		x       y
		-0.035976	16.34
		0.72755     25.41
		1.6605993	30.30
	    2.9658521	30.64	
	};
	\addplot+[sharp plot, solid, mark size=1.3pt, ultra thick] table {
		x       y
		0.678327	23.04
		1.087196	26.17
		1.8394697	29.16
		2.9658521	30.64
	};
	% agree
	\addplot+[sharp plot, solid,mark=triangle*,mark size=2pt, ultra thick] table {
		x       y
		0.6896127	24.44
		1.022566	27.38
		1.523082	29.60
	};
	
	\addplot[
	orange,
	ultra thick,
	visualization depends on=\thisrow{alignment} \as \alignment,
	nodes near coords,
	point meta=explicit symbolic,
	every node near coord/.style={anchor=\alignment} 
	] table [
	meta index=2 
	] {
		x       y         label       alignment
	    -1    31.36	offline   270
		5.794    31.36	offline   270
	};
	
	\addlegendentry{LA-2};    
	\addlegendentry{SP-2};    
	\addlegendentry{Hold-6}; 
	\addlegendentry{Offline};    

	% (Relative) Coordinate at top of the first plot
	\coordinate (c1) at (rel axis cs:0,1);
	%-----------
	% Plot [0, 1]
	%-----------

	\addplot[
	blue,
	only marks,
	mark=none,
	mark size=0.5pt,
	visualization depends on=\thisrow{alignment} \as \alignment,
	nodes near coords,
	point meta=explicit symbolic,
	every node near coord/.style={anchor=\alignment} 
	] table [
	meta index=2 
	] {
		x       y         label       alignment
		-0.035976	16.34   250  180
		0.22755     25.41   500 180
		1.5605993	30.30   1000   0
	    2.9658521	31.5   2000   0
	};
	\addplot[
	red,
	only marks,
	mark=none,
	mark size=0.5pt,
	visualization depends on=\thisrow{alignment} \as \alignment,
	nodes near coords,
	point meta=explicit symbolic,
	every node near coord/.style={anchor=\alignment} 
	] table [
	meta index=2 
	] {
		x       y         label       alignment
		0.678327	23.04   250 180
		1.087196	26.17   500 180
		1.8394697	28.7   1000    180
		3.1	29.5   2000    0
	};
	
	\addplot[
	brown,
	only marks,
	mark=none,
	mark size=0.5pt,
	visualization depends on=\thisrow{alignment} \as \alignment,
	nodes near coords,
	point meta=explicit symbolic,
	every node near coord/.style={anchor=\alignment} 
	] table [
	meta index=2 
	] {
		x       y         label       alignment
		0.6896127	24.44 500 180
		1.022566	27.38   1000    180 
		1.523082	29.60   2000    180
	};
	
	% (Relative) Coordinate at top of the second plot
%	\coordinate (c2) at (rel axis cs:1,1);
	\end{groupplot}
%	\coordinate (c3) at ($(c1)!.5!(c2)$);
	\node[above] at (4.8,0.25)
	{\pgfplotslegendfromname{legend_sizes}};
	\end{tikzpicture}%
	\caption{Quality-latency trade-off of different chunk sizes combined with different stable hypothesis detection strategies. The number next to the marks indicates chunk size in milliseconds. }
	\label{fig:chunk_sizes}
\end{figure}
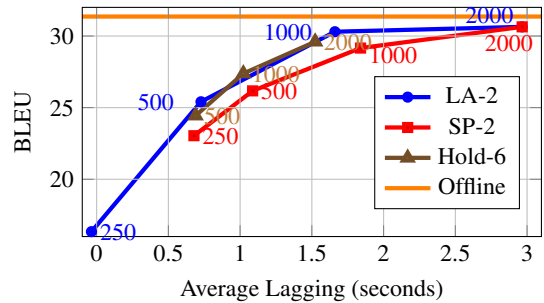

We experiment with chunk sizes of 250 ms, 500 ms, 1s, and 2 s. We combine the sizes of the chunks with different partial hypothesis selection strategies. The results are shown in \cref{fig:chunk_sizes}. 

The results document that the chunk size parameter has a stronger influence on the trade-off than different prefix strategies. Additionally, this enables constant trade-off strategies (e.g., LA-2) to become flexible.

\subsection{Stable Hypothesis Detection Strategies}
\label{sec:ex:strategies}
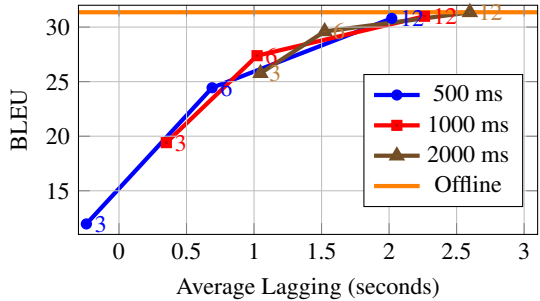
\begin{figure}[ht]
	\centering
	\footnotesize
	\begin{tikzpicture}
	\pgfplotsset{
		scaled y ticks = false,
		width=0.99\columnwidth,
		height=0.6\columnwidth,
		axis on top,
		grid=major,
		xlabel style={at={(0.5,0.1)}},
		extra x tick style={grid=none, draw=none, tick label style={xshift=0cm,yshift=.30cm},tick style={draw=none}},
		extra x tick label={$\mathbin{/\mkern-6mu/}$},
		ylabel shift={-1.5em},
		ylabel style={align=center, at={(0.1,0.5)}}
%		try min ticks=10
	}
	\begin{groupplot}[ 
	group style={
		group size=1 by 1,
		vertical sep=0pt,
		horizontal sep=0pt
	},
	]
	% ----------
	% Plot [0, 0]
	%-----------
	\nextgroupplot[
	xlabel={Average Lagging (seconds)},
	ylabel={BLEU},
    x label style={at={(axis description cs:0.5,0.02)},anchor=north},
    y label style={at={(axis description cs:0.11,.5)},anchor=south},
	xticklabels={0,0.5,1,1.5,2,2.5,3}, % Fake here the x-value for the offline thing
	xtick={0,0.5,1,1.5,2,2.5,3},
	yticklabels={15,20,25,30},
	ytick={15,20,25,30},
	xmin=-0.3,
	xmax=3.1,
	ymin=11, %13.9,
	ymax=32,
	legend style={legend columns=1,fill=white,draw=black,anchor=west,align=left, font=\footnotesize}, %at={($(0,0)+(1cm,1cm)$)},
	legend to name=legend_n
	]
	\addplot+[sharp plot, solid, mark size=1.3pt, ultra thick] table {
		x       y
		-0.24164	11.96
        0.6896127	24.44
        2.019868	30.80	
	};
	\addplot+[sharp plot, solid, mark size=1.3pt, ultra thick] table {
		x       y
		0.3505791	19.41
        1.022566	27.38
        2.2620401	30.99
	};
	% agree
	\addplot+[sharp plot, solid,mark=triangle*,mark size=2pt, ultra thick] table {
		x       y
		1.0476365	25.77
        1.523082	29.60
        2.598249	31.35
	};
	
	\addplot[
	orange,
	ultra thick,
	visualization depends on=\thisrow{alignment} \as \alignment,
	nodes near coords,
	point meta=explicit symbolic,
	every node near coord/.style={anchor=\alignment} 
	] table [
	meta index=2 
	] {
		x       y         label       alignment
	    -1    31.36	offline   270
		5.794    31.36	offline   270
	};
	
	\addlegendentry{500 ms};    
	\addlegendentry{1000 ms};    
	\addlegendentry{2000 ms}; 
	\addlegendentry{Offline};    

	% (Relative) Coordinate at top of the first plot
	\coordinate (c1) at (rel axis cs:0,1);
	%-----------
	% Plot [0, 1]
	%-----------

	\addplot[
	blue,
	only marks,
	mark=none,
	mark size=0.5pt,
	visualization depends on=\thisrow{alignment} \as \alignment,
	nodes near coords,
	point meta=explicit symbolic,
	every node near coord/.style={anchor=\alignment} 
	] table [
	meta index=2 
	] {
		x       y         label       alignment
		-0.24164	11.96   3   180
        0.6896127	24.44   6   180
        2.019868	30.80   12   180
	};
	\addplot[
	red,
	only marks,
	mark=none,
	mark size=0.5pt,
	visualization depends on=\thisrow{alignment} \as \alignment,
	nodes near coords,
	point meta=explicit symbolic,
	every node near coord/.style={anchor=\alignment} 
	] table [
	meta index=2 
	] {
		x       y         label       alignment
		0.3505791	19.41   3   180
        1.022566	27.38   6   180
        2.2620401	30.99   12   180
	};
	
	\addplot[
	brown,
	only marks,
	mark=none,
	mark size=0.5pt,
	visualization depends on=\thisrow{alignment} \as \alignment,
	nodes near coords,
	point meta=explicit symbolic,
	every node near coord/.style={anchor=\alignment} 
	] table [
	meta index=2 
	] {
		x       y         label       alignment
		1.0476365	25.77   3   180
        1.523082	29.60   6   180
        2.598249	31.35   12 180
	};
	
	% (Relative) Coordinate at top of the second plot
%	\coordinate (c2) at (rel axis cs:1,1);
	\end{groupplot}
%	\coordinate (c3) at ($(c1)!.5!(c2)$);
	\node[above] at (4.8,0.25)
	{\pgfplotslegendfromname{legend_n}};
	\end{tikzpicture}%
	\caption{Quality-latency trade-off of hold-$n$ strategy with different values of $n$. The number next to the marks indicates $n$. Colored lines connect results with equal chunk size.}
	\label{fig:hold_n}
\end{figure}

\begin{figure*}[ht]
	\centering
	\footnotesize
	\begin{tikzpicture}
	\pgfplotsset{
		scaled y ticks = false,
		width=1.8\columnwidth,
		height=1\columnwidth,
		axis on top,
		grid=major,
		xlabel style={at={(0.5,0.1)}},
		extra x tick style={grid=none, draw=none, tick label style={xshift=0cm,yshift=.30cm},tick style={draw=none}},
		extra x tick label={$\mathbin{/\mkern-6mu/}$},
		ylabel shift={-1.5em},
		ylabel style={align=center, at={(0.1,0.5)}}
%		try min ticks=10
	}
	\begin{groupplot}[ 
	group style={
		group size=1 by 1,
		vertical sep=0pt,
		horizontal sep=0pt
	},
	]
	% ----------
	% Plot [0, 0]
	%-----------
	\nextgroupplot[
	xlabel={Average Lagging (seconds)},
	ylabel={BLEU},
    x label style={at={(axis description cs:0.5,0.02)},anchor=north},
    y label style={at={(axis description cs:0.05,.5)},anchor=south},
	xticklabels={0,0.5,1,1.5,2,2.5,3}, % Fake here the x-value for the offline thing
	xtick={0,0.5,1,1.5,2,2.5,3},
	yticklabels={15,20,25,30},
	ytick={15,20,25,30},
	xmin=-0.3,
	xmax=3.1,
	ymin=11, %13.9,
	ymax=32,
	legend style={legend columns=1,fill=white,draw=black,anchor=west,align=left, font=\footnotesize}, %at={($(0,0)+(1cm,1cm)$)},
	legend to name=legend_la
	]
	\addplot+[sharp plot, mark size=1.3pt,  very thick] table {
		x       y
		-0.2730447	12.65
        -0.010811	13.76
        0.5073063	17.28
        1.0851314	22.63	
	};
	\addplot+[sharp plot, mark size=1.3pt,  very thick] table {
		x       y
		0.678327	23.04
1.087196	26.17
1.8394697	29.16
2.9658521	30.64
	};
	% agree
	\addplot+[sharp plot,mark=triangle*,mark size=2pt,  black!60!green, very thick] table {
		x       y
		-0.035976	16.34
0.72755	25.41
1.6605993	30.30
2.96656	31.41
	};
	
	\addplot+[sharp plot,mark=triangle*,mark size=2pt,  very thick] table {
		x       y
		0.6196735	24.55
1.45868555	29.72
2.61937445	31.21
4.09226348	31.44
	};
	
	\addplot+[sharp plot,mark=triangle*,mark size=2pt,  very thick] table {
		x       y
		1.034	27.92
        2.018	31.023
        3.334	31.43
        4.730	31.41
	};
	
    \addplot+[solid, sharp plot, mark size=1.3pt,  violet, very thick] table {
    		x       y
    		-0.24164	11.96
            0.6896127	24.44
            2.019868	30.80	
    	};
	
	\addplot[
	orange,
	ultra thick,
	visualization depends on=\thisrow{alignment} \as \alignment,
	nodes near coords,
	point meta=explicit symbolic,
	every node near coord/.style={anchor=\alignment} 
	] table [
	meta index=2 
	] {
		x       y         label       alignment
	    -1    31.36	offline   270
		5.794    31.36	offline   270
	};
	
	\addlegendentry{SP-1};    
	\addlegendentry{SP-2};    
	\addlegendentry{LA-2}; 
	\addlegendentry{LA-3}; 
	\addlegendentry{LA-4}; 
	\addlegendentry{Hold-$n$ 500}; 
	\addlegendentry{Offline}; 
%	\addlegendentry{Offline};    

	% (Relative) Coordinate at top of the first plot
	\coordinate (c1) at (rel axis cs:0,1);
	%-----------
	% Plot [0, 1]
	%-----------

	\addplot[
	blue,
	only marks,
	mark=none,
	mark size=0.5pt,
	visualization depends on=\thisrow{alignment} \as \alignment,
	nodes near coords,
	point meta=explicit symbolic,
	every node near coord/.style={anchor=\alignment} 
	] table [
	meta index=2 
	] {
		x       y         label       alignment
		-0.2730447	12.65	250	180
        -0.010811	13.76	500	180
        0.5073063	17.28	1000	180
        1.0851314	22.63	2000	180
	};
	\addplot[
	red,
	only marks,
	mark=none,
	mark size=0.5pt,
	visualization depends on=\thisrow{alignment} \as \alignment,
	nodes near coords,
	point meta=explicit symbolic,
	every node near coord/.style={anchor=\alignment} 
	] table [
	meta index=2 
	] {
		x       y         label       alignment
		0.678327	23.04	250	180
        1.087196	26.17	500	180
        1.8394697	29.16	1000	180
        2.9658521	30.64	2000	180
	};
	
	\addplot[
	black!60!green,
	only marks,
	mark=none,
	mark size=0.5pt,
	visualization depends on=\thisrow{alignment} \as \alignment,
	nodes near coords,
	point meta=explicit symbolic,
	every node near coord/.style={anchor=\alignment} 
	] table [
	meta index=2 
	] {
		x       y         label       alignment
		-0.035976	16.34	250	180
        0.72755	25.41	500	180
        1.6605993	30.30	1000	180
        2.96656	31.41	2000	180
	};

\addplot[
	black,
	only marks,
	mark=none,
	mark size=0.5pt,
	visualization depends on=\thisrow{alignment} \as \alignment,
	nodes near coords,
	point meta=explicit symbolic,
	every node near coord/.style={anchor=\alignment} 
	] table [
	meta index=2 
	] {
		x       y         label       alignment
        0.6196735	24.55	250	180
        1.45868555	29.72	500	180
        2.61937445	31.21	1000	180
        4.09226348	31.44	2000	180
	};
	
	\addplot[
	blue,
	only marks,
	mark=none,
	mark size=0.5pt,
	visualization depends on=\thisrow{alignment} \as \alignment,
	nodes near coords,
	point meta=explicit symbolic,
	every node near coord/.style={anchor=\alignment} 
	] table [
	meta index=2 
	] {
		x       y         label       alignment
        1.034	27.92   250 180
        2.018	31.023 500  180
        3.334	31.43   1000    180 
        4.730	31.41   2000    180
	};
	
	\addplot[
	purple,
	only marks,
	mark=none,
	mark size=0.5pt,
	visualization depends on=\thisrow{alignment} \as \alignment,
	nodes near coords,
	point meta=explicit symbolic,
	every node near coord/.style={anchor=\alignment} 
	] table [
	meta index=2 
	] {
		x       y         label       alignment
		-0.24164	11.96   3   180
        0.6896127	24.44   6   180
        2.019868	30.80   12   180
	};

	% (Relative) Coordinate at top of the second plot
%	\coordinate (c2) at (rel axis cs:1,1);
	\end{groupplot}
%	\coordinate (c3) at ($(c1)!.5!(c2)$);
	\node[above] at (10.8,0.25)
	{\pgfplotslegendfromname{legend_la}};
	\end{tikzpicture}%
	\caption{Quality-latency trade-off of shared prefix (SP-$n$) and local agreement (LA-$n$) with different $n$ and chunk size.}
	\label{fig:la}
\end{figure*}
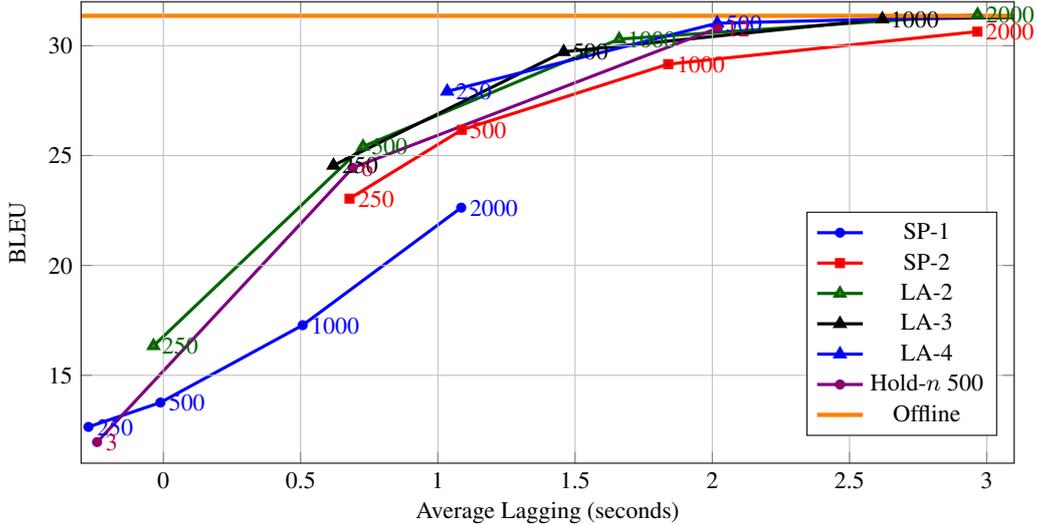

We experiment with three strategies: hold-$n$ (withholds last $n$ tokens), shared prefix (SP-$n$; finds the longest common prefix of all beams in $n$ consecutive chunks) and local agreement (LA-$n$; finds the longest common prefix of the best hypothesis in $n$ consecutive chunks). For hold-$n$, we select $n=3,6,12$; for SP-$n$, we select $n=1,2$ ($n=1$ corresponds to the strategy by \citet{nguyen21c_interspeech}); for LA-$n$ we select $n=2,3,4$ ($n=2$ corresponds to the strategy by \citet{Liu2020}). The results are in \cref{fig:hold_n,fig:la}.


\paragraph{Hold-$n$}
The results suggest (see \cref{fig:hold_n}) that the hold-$n$ strategy can use either $n$ or chunk size to control the quality-latency trade-off with equal effect. The only exception seems to be too low $n <= 3$, which slightly underperforms the options with higher $n$ and shorter chunk size.

\paragraph{Local agreement (LA-$n$)}
The local agreement seems to outperform all other strategies (see \cref{fig:la}). LA-$n$ for all $n$ follows the same quality-latency trade-off line. The advantage of LA-2 is in reduced computational complexity compared to the other LA-$n$ strategies with $n > 2$.

\paragraph{Shared prefix (SP-$n$)}
SP-1 strongly underperforms other strategies in quality (see \cref{fig:la}). While the SP-1 strategy performs well in the ASR task \citep{nguyen21c_interspeech}, it is probably too lax for the speech translation task. The generalized and more conservative SP-2 performs much better. Although, the more relaxed LA-2, which considers only the best item in the beam, has a better quality-latency trade-off curve than the more conservative SP-2.

\subsection{Initial Wait}
As we could see in \cref{exp:chunks}, the shorter chunk sizes tend to perform worse. One of the reasons might be the limited context of the early chunks.\footnote{If we translated a non-pre-segmented input, this problem would be limited only onetime to the beginning of the input.} To increase the early context, we prolong the first chunk to 2 seconds.

\begin{table}[ht]
\centering
\resizebox{0.45\textwidth}{!}{%
\begin{tabular}{@{}clrrrr@{}}
\toprule
\multicolumn{1}{l}{Initial wait} & \multicolumn{1}{l|}{Chunk size}                & BLEU  & AL      & AP   & DAL     \\ \midrule
\multirow{3}{*}{0}               & \multicolumn{1}{l|}{250}  & 16.34 & -35.97  & 0.66 & 1435.06 \\
                                 & \multicolumn{1}{l|}{500}  & 25.40 & 727.55  & 0.73 & 1791.21 \\
                                 & \multicolumn{1}{l|}{1000}                      & 30.29 & 1660.59 & 0.83 & 2662.18 \\ \midrule
\multirow{3}{*}{2000}            & \multicolumn{1}{l|}{250}  & 16.60 & 358.35  & 0.74 & 2121.54 \\
                                 & \multicolumn{1}{l|}{500}  & 25.42 & 952.15  & 0.77 & 2142.53 \\
                                 & \multicolumn{1}{l|}{1000} & 30.29 & 1654.77 & 0.83 & 2657.48 \\ \midrule
\end{tabular}%
}
\caption{Quality-latency trade-off of the LA-2 strategy with and without the initial wait.}
\label{tab:initial_wait}
\end{table}

The results are in \cref{tab:initial_wait}. We see a slight (0.3 BLEU) increase in quality for a chunk size of 250 ms, though the initial wait does not improve the BLEU and a considerable increase in the latency.  

The performance seems to be influenced mainly by the chunk size. The reason for smaller chunks' under-performance might be caused by (1) acoustic uncertainty towards the end of a chunk (e.g., words often get cut in the middle), or (2) insufficient information difference between two consecutive chunks. 

This is supported by the observation in \cref{fig:la}. Increasing the number of consecutive chunks (i.e., increasing the context for the decision) considered in the local agreement strategy (LA-$2,3,4$), improves the quality, while it adds latency. 

\subsection{Negative Average Lagging}
Interestingly, we noticed that some of the strategies achieved negative average lagging (e.g., LA-2 in \cref{exp:chunks}) with a chunk size of 250 ms has AL of -36 ms). After a closer examination of the outputs, we found that the negative AL is in utterances where the hypothesis is significantly longer than the reference. Recall the average latency for speech input defined by \citet{ma2020simuleval}:
\begin{equation}
    \text{AL}_{\text{speech}} = \frac{1}{\tau'(|\mathbf{X}|)} \sum_{i=1}^{\tau'(|\mathbf{X}|)} d_i - d^*_i,
    \label{eq:al_speech}
\end{equation}
where $d_i = \sum_{k=1}^j T_k$, $j$ is the index of raw audio segment that has been read when generating $y_i$, $T_k$ is duration of raw audio segment, $\tau'(|\mathbf{X}|) = \text{min}\{i | d_i=\sum_{j=1}^{|\mathbf{X}|}T_j\}$ and $d^*_i$ are the delays of an ideal policy:
\begin{equation}
    d^*_i = (i-1)\times\sum_{j=1}^{|\mathbf{X}|}T_j\ / |\mathbf{Y^*}|,
    \label{eq:ideal}
\end{equation}
where $\mathbf{Y^*}$ is reference translation. 

If the hypothesis is longer than the reference, then $d^*_i > d_i$, making the sum argument in \cref{eq:al_speech} negative. On the other hand, if we use the length of the hypothesis instead, then a shorter hypothesis would benefit.\footnote{\citet{ma2019stacl} originally used the hypothesis length in the \cref{eq:ideal} and then \citet{ma2020simuleval} suggested to use the reference length instead.} We, therefore, suggest using the maximum of both to prevent the advantage of either a shorter or a longer hypothesis:
\begin{equation}
    d^*_i = (i-1)\times\sum_{j=1}^{|\mathbf{X}|}T_j\ / max(|\mathbf{Y}|,|\mathbf{Y^*}|).
    \label{eq:ideal_new}
\end{equation}

\section{Submitted System}
\label{sec:submitted}
In this section, we describe the submitted system. We follow the allowed training data and pretrained models and therefore our submission is \emph{constrained} (see \cref{sec:models} for model description).

\begin{figure}[ht]
	\centering
	\footnotesize
	\begin{tikzpicture}
	\pgfplotsset{
		scaled y ticks = false,
		width=0.99\columnwidth,
		height=0.6\columnwidth,
		axis on top,
		grid=major,
		xlabel style={at={(0.5,0.1)}},
		extra x tick style={grid=none, draw=none, tick label style={xshift=0cm,yshift=.30cm},tick style={draw=none}},
		extra x tick label={$\mathbin{/\mkern-6mu/}$},
		ylabel shift={-1.5em},
		ylabel style={align=center, at={(0.1,0.5)}}
%		try min ticks=10
	}
	\begin{groupplot}[ 
	group style={
		group size=1 by 1,
		vertical sep=0pt,
		horizontal sep=0pt
	},
	]
	% ----------
	% Plot [0, 0]
	%-----------
	\nextgroupplot[
	xlabel={Average Lagging (seconds)},
	ylabel={BLEU},
    x label style={at={(axis description cs:0.5,0.02)},anchor=north},
    y label style={at={(axis description cs:0.11,.5)},anchor=south},
	xticklabels={0,0.5,1,1.5,2,2.5,3,3.5}, % Fake here the x-value for the offline thing
	xtick={0,0.5,1,1.5,2,2.5,3,3.5},
	yticklabels={28,30,32},
	ytick={28,30,32},
	xmin=0.8,
	xmax=3.8,
	ymin=26.5, %13.9,
	ymax=33.5,
	legend style={legend columns=1,fill=white,draw=black,anchor=west,align=left, font=\footnotesize}, %at={($(0,0)+(1cm,1cm)$)},
	legend to name=legend_sizes_compare
	]
	\addplot+[sharp plot, solid, mark size=1.3pt, ultra thick] table {
		x       y
		0.947	27.05
		1.660	30.30
		2.966	31.41
	};
	
	% agree
	\addplot+[sharp plot, solid,mark=triangle*,mark size=2pt, ultra thick] table {
		x       y
		0.947	26.93
		1.906	31.60
		3.663	32.98
	};
	\addplot+[sharp plot, solid, mark size=1.3pt, ultra thick] table {
		x       y
		0.920	 27.40
		1.860     29.68
		2.740	30.75
	};

	\addlegendentry{Model A};    
	\addlegendentry{Model B}; 
	\addlegendentry{Best IWSLT21};    

	% (Relative) Coordinate at top of the first plot
	\coordinate (c1) at (rel axis cs:0,1);
	%-----------
	% Plot [0, 1]
	%-----------

	% (Relative) Coordinate at top of the second plot
%	\coordinate (c2) at (rel axis cs:1,1);
	\end{groupplot}
%	\coordinate (c3) at ($(c1)!.5!(c2)$);
	\node[above] at (4.5,0.0)
	{\pgfplotslegendfromname{legend_sizes_compare}};
	\end{tikzpicture}%
	\caption{Quality-latency trade-off on English-German tst-COMMON of our two models: a dedicated English-German model trained from scratch (Model A) and a multilingual model based on wav2vec and mBART (Model B). We also include the best IWSLT 2021 system (USTC-NELSLIP \citep{liu-etal-2021-ustc}). }
	\label{fig:compare}
\end{figure}
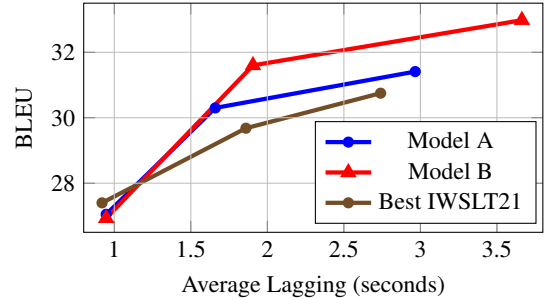

For stable hypothesis detection, we decided to use the local agreement strategy with $n=2$. As shown in \cref{sec:ex:strategies}, the LA-2 has the best latency-quality trade-off along with other LA-$n$ strategies. To achieve the different latency regimes, we use various chunk sizes, depending on the language pair. We decided not to use larger $n>2$ to control the latency, as it increases the computation complexity while having the same effect as using a different chunk size. The results on MuST-C tst-COMMON are in \cref{tab:submission}. The quality-latency trade-off is in \cref{fig:compare}. 

% \usepackage{multirow}
% \usepackage{graphicx}
\begin{table*}[ht]
\centering
\resizebox{.8\textwidth}{!}{%
\begin{tabular}{l|clc|rrrr}
\hline
Model & \multicolumn{1}{l}{Language pair} & Latency regime & \multicolumn{1}{l|}{Chunk size} & BLEU & AL & AP & DAL \\ \hline
\multirow{3}{*}{Best IWSLT21 system}  
                          & \multirow{3}{*}{En-De} & Low     & - & 27.40 & 920  & 0.68 & 1420 \\
                          &                        & Medium  & - & 29.68 & 1860 & 0.82 & 2650 \\
                          &                        & High    & - & 30.75 & 2740 & 0.90 & 3630 \\ \hline
\multirow{4}{*}{Model A}  & \multirow{4}{*}{En-De} & Low     & 600  & 27.05 & 947 & 0.76 & 1993 \\
                          &                        & Medium  & 1000 & 30.30 & 1660 & 0.84 & 2662 \\
                          &                        & High    & 2000 & 31.41 & 2966 & 0.93 & 3853 \\
                          &                        & Offline & -    & 31.36 & 5794 & 1.00 & 5794 \\ \hline
\multirow{12}{*}{Model B} & \multirow{4}{*}{En-De} & Low     & 500  & 26.93 & 945 & 0.77 & 2052 \\
                          &                        & Medium  & 1000 & 31.60 & 1906 & 0.86 & 2945 \\
                          &                        & High    & 2500 & 32.98 & 3663 & 0.96 & 4452 \\
                          &                        & Offline & -    & 33.14 & 5794 & 1.00 & 5794 \\ \cline{2-8} 
                          & \multirow{4}{*}{En-Ja} & Low     & 1000 & 16.84 & 2452 & 0.90 & 3212 \\
                          &                        & Medium  & 2400 & 16.99 & 3791 & 0.97 & 4296 \\
                          &                        & High    & 3000 & 16.97 & 4140 & 0.98 & 4536 \\
                          &                        & Offline & -    & 16.88 & 5119 & 1.00 & 5119 \\ \cline{2-8} 
                          & \multirow{4}{*}{En-Zh} & Low     & 800  & 23.69 & 1761 & 0.85 & 2561 \\
                          &                        & Medium  & 1500 & 24.29 & 2788 & 0.93 & 3500 \\
                          &                        & High    & 2500 & 24.56 & 3669 & 0.97 & 4212 \\
                          &                        & Offline & -    & 24.54 & 5119 & 1.00 & 5119 \\ \hline
\end{tabular}%
}
\caption{Results of the older model used for the experiments (Model A) and the submitted system (Model B) on the MuST-C v2 tst-COMMON. We also include the best IWSLT 2021 system (USTC-NELSLIP \citep{liu-etal-2021-ustc}).}
\label{tab:submission}
\end{table*}

From \cref{tab:submission,fig:compare}, we can see that the proposed method works well on two different models and various language pairs. We see that an improvement in the offline model (offline BLEU of 31.36 and 33.14 for Model A and B, respectively) leads to improvement in the online regime. 

Finally, we see that our method beats the best IWSLT 2021 system (USTC-NELSLIP  \citep{liu-etal-2021-ustc}) in medium and high latency regimes using both models (i.e., a model trained from scratch and a model based on pretrained wav2vec and mBART), and is almost on par in the low latency regime (Model A is losing 0.35 BLEU and Model B is losing 0.47 BLEU).

\subsection{Computationally Aware Latency}
In this paper, we do not report any computationally aware metrics, as our implementation of Transformers is slow. Later, we implemented the same onlinization approach using wav2vec 2.0 and mBART from Huggingface Transformers \cite{wolf-etal-2020-transformers}. The new implementation reaches faster than real-time inference speed.

\section{Conclusion}
In this paper, we reviewed onlinization strategies for end-to-end speech translation models. We identified the optimal stable hypothesis detection strategy and proposed two separate ways of the quality-latency trade-off parametrization. We showed that the onlinization of the offline models is easy and performs almost on par with the offline run. We demonstrated that an improvement in the offline model leads to improved online performance. We also showed that our method outperforms a dedicated simultaneous system. Finally, we proposed an improvement in the average latency metric.

\section*{Acknowledgments}
This work has received support from the project ``Grant Schemes at CU'' (reg. no. CZ.02.2.69/0.0/0.0/19\_073/0016935), %
the grant 19-26934X (NEUREM3) of the Czech Science Foundation, %
the European Union's Horizon 2020 Research and Innovation Programme under Grant Agreement No 825460 (ELITR), %
and partly supported by a Facebook Sponsored Research Agreement ``Language Similarity in Machine Translation''. %

\bibliography{anthology,custom}
\bibliographystyle{acl_natbib}

\end{document}